\documentclass[letterpaper, 10 pt, conference]{ieeeconf}  

\IEEEoverridecommandlockouts                              

\usepackage{amsmath}
\usepackage{amssymb}
\usepackage{amsfonts}
\usepackage{latexsym}
\usepackage{lipsum}
\usepackage{multicol}
\usepackage{accents}
\usepackage{harpoon}
\usepackage{wrapfig}
\usepackage{tabularx}
\usepackage{euscript}
\usepackage{color}
\usepackage{xcolor}
\usepackage{graphicx}
\usepackage{float}
\usepackage[caption = false]{subfig}	
\usepackage{epsfig}
\usepackage{hyperref}
\usepackage{setspace}
\usepackage{fancyhdr}
\usepackage{textcomp}
\usepackage{cite}
\usepackage{epstopdf}
\usepackage{subfig}
\usepackage{verbatim}
\DeclareGraphicsExtensions{.pdf,.eps,.jpg}
\graphicspath{{Figures/}}

\usepackage{pifont}
\usepackage{latexsym}
\usepackage{psfrag}
\usepackage{stmaryrd}

\usepackage[linesnumbered,ruled,vlined]{algorithm2e}

\title{\LARGE \bf
Cooperative and Distributed Reinforcement Learning of Drones for Field Coverage
}

\author{Huy Xuan Pham, Hung Manh La, David Feil-Seifer, and Ara Nefian
\thanks{This material is based upon work supported by the National Aeronautics and Space Administration (NASA) Grant No. NNX15AK48A issued through the NV EPSCoR RID Seed, and NASA Grant No. NNX15AI02H issued through the NVSGC Faculty Award - CD.}
\thanks{Huy Pham is a PhD student of the Advanced Robotics and Automation
(ARA) Laboratory, and Dr. Hung La is the director of ARA Lab. Dr. David Feil-Seifer is an Assistant Professor of Department of Computer Science and Engineering, University
of Nevada, Reno, NV 89557, USA. Dr. Ara Nefian is with NASA Ames Research Center, Moffett Field, CA 94035.
Corresponding author: Hung La, email: {\tt\small hla@unr.edu}}
}

\begin{document}

\maketitle
\thispagestyle{empty}
\pagestyle{empty}

\begin{abstract}
This paper proposes a distributed Multi-Agent Reinforcement Learning (MARL) algorithm for a team of Unmanned Aerial Vehicles (UAVs). The proposed MARL algorithm allows UAVs to learn cooperatively to provide a full coverage of an unknown field of interest while minimizing the overlapping sections among their field of views. Two challenges in MARL for such a system are discussed in the paper: firstly, the complex dynamic of the joint-actions of the UAV team, that will be solved using game-theoretic correlated equilibrium, and secondly, the challenge in huge dimensional state space representation will be tackled with efficient function approximation techniques. We also provide our experimental results in detail with both simulation and physical implementation to show that the UAV team can successfully learn to accomplish the task.
\end{abstract}


\section{Introduction}\label{S.intro}

Optimal sensing coverage is an active research branch. Solutions have been proposed in previous work, for instance, by solving general locational optimization problem~\cite{cortes2004coverage}, using Voronoi partitions~\cite{schwager2009decentralized, breitenmoser2010voronoi}, using potential field methods~\cite{ge2002dynamic, schwager2011eyes}, or scalar field mapping~\cite{la2013distributed,MNguyen_TCNS2017}. In most of those work, authors made assumption about the mathematical model of the environment, such as distribution model of the field or the predefined coverage path~\cite{la2015multi, la2015cooperative}. In reality,  however, it is very difficult to have an accurate model, because its data is normally limited or unavailable.

Unmanned aerial vehicles (UAV), or drones, have already become popular in human society, with a wide range of application, from retailing business to environmental issues. The ability to provide visual information with low costs and high flexibility makes the drones  preferable equipment in tasks relating to field coverage and monitoring, such as in wildfire monitoring~\cite{pham2017distributed}, or search and rescue~\cite{tomic2012toward}. In such applications, usually a team of UAVs could be deployed to increase the coverage range and reliability of the mission. As with other multi-agent systems \cite{TNguyen_TCNS2017, LA_RAS2012}, the important challenges in designing an autonomous team of UAVs for field coverage include dealing with the dynamic complexity of the interaction between the UAVs so that they can coordinate to accomplish a common team goal. 

Model-free learning algorithms, such as Reinforcement learning (RL), would be a natural approach to address the aforementioned challenges relating to the required accurate mathematical models for the environment, and the complex behaviors of the system. These algorithms will allow each agent in the team to learn new behavior, or reach consensus with others  \cite{RNC:RNC3687}, without depending on a model of the environment~\cite{sutton1998reinforcement}. Among them, RL is popular because it is relatively generic to address a wide range of problem, while it is simple to implement.

Classic individual RL algorithms have already been extensively researched in UAV applications. Previous papers focus on applying RL algorithm into UAV control to achieve desired trajectory tracking/following~\cite{bou2010controller}, or discussion of using RL to improve the performance in UAV application~\cite{faust2013learning}. Multi-Agent Reinforcement Learning (MARL) is also an active field of research. In multi-agent systems (MAS), agents' behaviors cannot be fully designed as \textit{priori} due to the complicated nature, therefore, the ability to learn appropriate behaviors and interactions will provide a huge advantage for the system. This particularly benefits the system when new agents are introduced, or the environment is changed~\cite{bucsoniu2010multi}. Recent publications concerning the possibility of applying MARL into a variety of applications, such as in autonomous driving~\cite{shalev2016safe}, or traffic control~\cite{bakker2010traffic}. 

In robotics, efforts have been focused on robotic system coordination and collaboration~\cite{ma2008combining}, transfer learning~\cite{helwa2017multi}, or multi-target observation~\cite{fernandez2001learning}. For robot path planning and control, most prior research focuses on classic problems, such as navigation and collision avoidance~\cite{hu2003nash}, object carrying by robot teams~\cite{sadhu2017improving}, or pursuing preys/avoiding predators~\cite{ishiwaka2003approach, la2015multirobot}. Many other papers in multi-robotic systems even simplified the dynamic nature of the system to use individual agent learning such as classic RL algorithm~\cite{hung2017q}, or actor-critic model~\cite{adepegba2016multi}. To our best knowledge, not so many works available addressed the complexity of MARL in a multi-UAV system and their daily missions such as optimal sensing coverage. In this paper, we propose how a MARL algorithm can be applied to solve an optimal coverage problem. We address two challenges in MARL: (1) the complex dynamic of the joint-actions of the UAV team, that will be solved using game-theoric correlated equilibrium, and (2) the challenge in huge dimensional state space will be tackled with an efficient space-reduced representation of the value function. 

The remaining of the paper is organized as follows. Section \ref{S.2} details on the optimal field coverage problem formulation. In section \ref{S.3}, we discuss our approach to solve the problem and the design of the learning algorithm. Basics in MARL will also be covered. We present our experimental result in section \ref{S.4} with a comprehensive simulation, followed by an implementation with physical UAVs in a lab setting. Finally, section \ref{S.5} concludes our paper and layouts future work.

\section{Problem Formulation}\label{S.2}
\begin{figure}[htb]
\centering
\includegraphics[width=1\columnwidth]{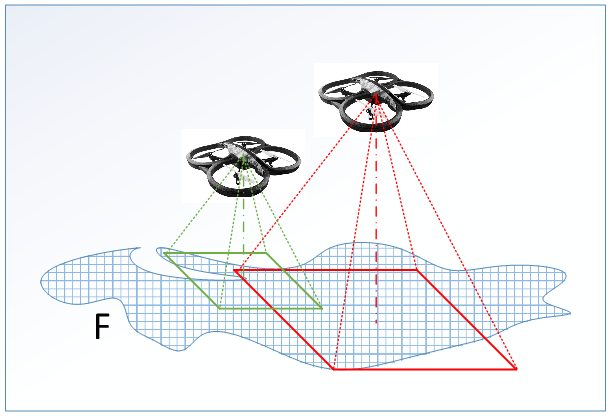}
  \caption{A team of UAVs to cover a field of interest $F$. A UAV can enlarge the FOV by flying higher, but risk in getting overlapped with other UAVs in the system. Minimizing overlap will increase the field coverage and resolution.}
  \label{F.Problem}
\vspace{-5 pt}
\end{figure}
\begin{figure}[htb!]
\centering
\includegraphics[width=0.9\columnwidth]{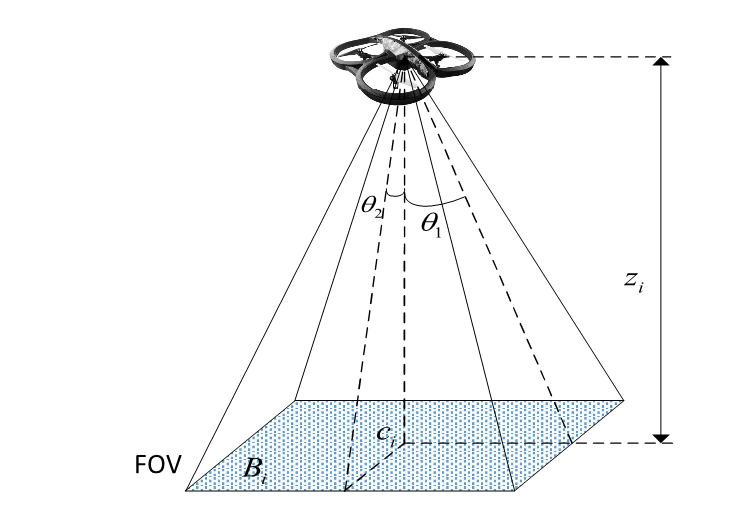}
  \caption{Field of view of each UAV.}
  \label{F.FOV}
\vspace{-10 pt}
\end{figure}

In an mission like exploring a new environment such as monitoring an oil spilling or an wildfire area, it is growing interest to send out a fleet of UAVs acting as a mobile sensor network, as it provides many advantages comparing to traditional static monitoring methods~\cite{la2013distributed}. In such mission, the UAV team needs to surround the field of interest to get more information, for example, visual data. Suppose that we have a team of quadrotor-type UAVs (Figure \ref{F.Problem}). Each UAV is an independent decision maker, thus the system is distributed. They can localize itself using on-board localization system, such as using GPS. They can also exchange information with other UAVs through communication links. Each UAV equipped with identical downward facing cameras provides it a square field of view (FOV). The camera of each UAV and its FOV form a pyramid with half-angles $\theta^{T} =[\theta_{1}, \theta_{2}]^{T}$ (Figure \ref{F.FOV}). A point $q$ is covered by the FOV of UAV $i$ if it satisfies the following equations:
\begin{equation}\label{2.coveragecondition}
\begin{aligned}
	\frac{||q - c_{i}||}{z_{i}} \leq \tan \theta^{T}, \\
\end{aligned}
\end{equation}
\noindent
\noindent
where $c_{i}$ is the lateral-projected position, and $z_{i}$ is the altitude of the UAV $i$, respectively. The objective of the team is not only to provide a full coverage over the shape of the field of interest $F$ under their UAVs' FOV, but also to minimize overlapping other UAVs' FOV to improve the efficiency of the team (e.g., minimizing overlap can increase resolution of field coverage). A UAV covers $F$ by trying to put a section of it under its FOV. It can enlarge the FOV to cover a larger section by increasing the altitude $z_{i}$ according to (\ref{2.coveragecondition}), however it may risk overlapping other UAVs' FOV in doing so. Formally speaking, let us consider a field $F$ of arbitrarily shapes. Let $p_{1},p_{2},...,p_{m}$ denote the positions of $m$ UAV $1, 2, ..., m$, respectively. Each UAV $i$ has a square FOV projected on the environment plane, denoted as $B_{i}$. Let $f(q,p_{1},p_{2},...,p_{m})$ represents a combined areas under the FOVs of the UAVs. The team has a cost function $H$ represented by:
\begin{equation}\label{2.Objective}
\begin{aligned}
	\ H &= \int_{q\in F}f(q,p_{1},p_{2},...,p_{m})\Phi(q)dq \\
		& - \int_{q\in B_{i} \cap B_{j}, \forall i, j \in m}f(q,p_{1},p_{2},...,p_{m})\Phi(q)dq, \\
\end{aligned}
\end{equation}
\noindent
where $\Phi(q)$ measures the importance of a specific area. In a plain field of interest, $\Phi(q)$ is constant. 

The problem can be solved using traditional methods, such as, using Voronoi partitions~\cite{schwager2009decentralized, breitenmoser2010voronoi}, or using potential field methods~\cite{ge2002dynamic, schwager2011eyes}. Most of these works proposed model-based approach, where authors made assumption about the mathematical model of the environment, such as the shape of the target~\cite{la2015multi, la2015cooperative}. In reality, however, it is very difficult to obtain an accurate model, because the data of the environment is normally insufficiently or unavailable. This can be problematic, as the systems may fail if using incorrect models. On the other hand, many learning algorithm, such as RL algorithms, rely only on the data obtained directly from the system, would be a natural option to address the problem.

\section{Algorithm}\label{S.3}
\subsection{Reinforcement Learning and Multi-Agent Reinforcement Learning}

Classic RL defines the learning process happens when a decision maker, or an agent, interacts with the environment. During the learning process, the agent will select the appropriate actions when presented a situation at each state according to a policy $\pi$, to maximize a numerical reward signal, that measures the performance of the agent, feedback from the environment. In MAS, the agents interact with not only the environment, but also with other agents in the system, making their interactions more complex. The state transition of the system is more complicated, resulting from a join action containing all the actions of all agents taking at a time step. The agents in the system now must also consider other agents states and actions to coordinate and/or compete with. Assuming that the environment has Markovian property, where the next state and reward of an agent only depends on the current state, the Multi-Agent Learning model can be generalized as a Markov game $<m, \{S\}, \{A\}, T, R>$, where:
\begin{itemize}
	\item $m$ is the number of agents in the system.
	\item $\{S\}$ is the joint state space $\{S\} = \bold{S_{1}} \times \bold{S_{2}} \times ... \times \bold{S_{m}}$, where $\bold{S_{i}}, i = 1, ...,m$ is the individual state space of an agent $i$. At time step $k$, the individual state of agent $i$ is denoted as $s_{i, k}$. The joint state at time step $k$ is denoted as $S_{k} = \{s_{1, k}, s_{2, k}, ..., s_{m, k}\}$.
	\item $\{A\}$ is the joint action space, $\{A\} = \bold{A_{1}} \times \bold{A_{2}} \times ... \times \bold{A_{m}}$, where $\bold{A_{i}}, i = 1, ...,m$ is the individual action space of an agent $i$. Each joint action at time $k$ is denoted as $A_{k} \in \{A\}$ while the individual action of agent $i$ is denoted as $a_{i, k}$. We have: $A_{k} = \{a_{1, k}, a_{2, k}, ..., a_{m, k}\}$.
	\item $T$ is the transition probability function, $T:S \times \bold{A} \times \bold{S}  \rightarrow [0, 1]$, is the probability of agent $i$ that takes action $a_{i,k}$ to move from state $s_{i,k}$ to state $s_{i,k+1}$. Generally, it is represented by a probability: $T(s_{i,k}, a_{i,k}) =  P(s_{i,k+1}|s_{i,k}, a_{i,k}) = P_{i}(a_{k})$.
	\item $R$ is the individual reward function: $R: \bold{S} \times \bold{A} \rightarrow \mathbb{R}$ that specifies the immediate reward of the agent $i$ for getting from $s_{i,k}$  at time step $k$ to state $s_{i, k+1}$ at time step $k+1$ after taking action $a_{i,k}$. In MARL, the team has a global reward $GR: \{S\} \times \{A\} \rightarrow \mathbb{R}$ in achieving the team's objective. We have: $GR(S_{k}, A_{k}) = r_{k+1}$.
\end{itemize}

The agents seek to optimize expected rewards in an \textit{episode} by determining which action to take that will have the highest return in the long run. In single agent learning, a value function $Q(s_{k}, a_{k})$, $\bold{A} \times \bold{S}  \rightarrow \mathbb{R}$, helps quantify strategically how good the agent will be if it takes an action $a_{k}$ at state $s_{k}$, by calculating its expected return obtained over an episode. In MARL, the action-state value function of each agent also depends on the joint state and joint action~\cite{nowe2012game}, represented as:
\begin{equation}\label{2.ValueFunctionJoint}
\begin{aligned}
	Q(s_{i, k}, a_{i, k}, s_{-i,k}, a_{-i,k}) = Q(S_{k}, A_{k}) = E\{ \sum_{k}^{\infty}\gamma r_{i, k+1} \},\\
\end{aligned}
\end{equation}
\noindent
where $0 < \gamma \leq 1$ is the discount factor of the learning. This function is also called Q-function. It is obvious that the state space and action space, as well as the value function in MARL is much larger than in individual RL, therefore MARL would require much larger memory space, that will be a huge challenge concerning the scalability of the problem.

\subsection{Correlated Equilibrium}

In order to accomplish the team's goal, the agents must reach consensus in selecting actions. The set of actions that they agreed to choose is called a joint action, $A_{k} \in \{A\}$. Such an agreement can be evaluated at equilibrium, such as Nash equilibrium (NE)~\cite{hu2003nash} or Correlated equilibrium (CE)~\cite{greenwald2003correlated}. Unlike NE, CE can be solved with the help of linear programming (LP)~\cite{papadimitriou2008computing}. Inspired by~\cite{greenwald2003correlated}  and~\cite{sadhu2017improving}, in this work we use a strategy that computes the optimal policy by finding the CE equilibrium for the agents in the systems. From the general problem of finding CE in game theory~\cite{papadimitriou2008computing}, we formulate a LP to help find the stable action for each agent as follows:
\begin{equation}\label{2.CE_pure}
\begin{aligned}
	&\pi(A_{k}) = \underset{A_{k}}{\arg \max}\{\sum_{i = 1}^{m}Q_{i,k}(S_{k}, A_{k}))P_{i}(a_{k}) \}.\\
	 &\text{subject to: } \\	 
	\ &\sum_{a_{k} \in \bold{A_{i}}}P_{i}(a_{k}) = 1, \forall i \in \{ m \}\\
	\ &P_{i}(a_{k}) \geq 0, \forall i \in \{ m \}, \forall a_{k} \in \bold{A_{i}} \\
	\ &\sum_{a'_{k} \in \bold{A_{i}}}[Q_{i,k}(S_{k},a_{k},A_{k,-i}) - Q_{i,k}(S_{k},a'_{k},A_{k,-i})]P_{i}(a_{k}) \\
	\ &\geq 0, \forall i \in \{ m \}.
\end{aligned}
\end{equation}
\noindent
Here, $P_{i}(a_{k})$ is the probability of UAV $i$ selecting action $a$ at time $k$, and $A_{-i}$ denotes  the rest of the actions of other agents. Solving LP has long been researched by the optimization community. In this work, we use a state-of-the-art program from the community to help us solve the above LP.
\subsection{Learning Design}

In this section, we design a MARL algorithm to solve our problem formulated in section \ref{S.2}. We assume that the system is fully observable. We also assume the UAVs are identical, and operated in the same environment, and have identical sets of states and actions: $\bold{S_{1}}= \bold{S_{2}}= ... = \bold{S_{m}}$, and $\bold{A_{1}}= \bold{A_{2}}= ... = \bold{A_{m}}$.

The state space and action space set of each agent should be represented as discrete finite sets approximately, to guarantee the convergence of the RL algorithm~\cite{busoniu2010reinforcement}. We consider the environment as a 3-D grid, containing a finite set of cubes, with the center of each cube represents a discrete location of the environment. The state of an UAV $i$ is defined as its approximate position in the environment, $s_{i, k} \triangleq [x_{c}, y_{c}, z_{c}] \in \bold{S}$, where $x_{c}$, $y_{c}$, $z_{c}$ are the coordinates of the center of a cube $c$ at time step $k$. The objective equation (\ref{2.Objective}) now becomes:
\begin{equation}\label{2.DiscreteObjective}
\begin{aligned}
	\underset{S_{k} \in \{S\}}{\max H} = \underset{S_{k} \in \{S\}}{\arg \max}\{\sum_{i}f_{i}(S_{k}) - \sum_{i}o_{i}(S_{k})\}, \\
\end{aligned}
\end{equation}
\noindent
where $f_{i}: \{S\} \rightarrow \mathbb{R}$ is the count of squares, or cells, approximating the field $F$ under the FOV of UAV $i$, and $o_{i}: \{S\} \rightarrow \mathbb{R}$ is the total number of cells overlapped with other UAVs. 

To navigate, each UAV $i$ can take an action $a_{i, k}$ out of a set of six possible actions $A$: heading North, West, South or East in lateral direction, or go Up or Down to change the altitude. Note that if the UAV stays in a state near the border of the environment, and selects an action that takes it out of the space, it should stay still in the current state. Certainly, the action $a_{i, k}$ belongs to an optimal joint-action strategy $A_{k}$ resulted from (\ref{2.CE_pure}). Note that in case multiple equilibrium exists, since each UAV is an independent agent, they can choose different equilibrium, making their respective actions deviate from the optimal joint action to a sub-optimal joint action. To overcome this, we employ a mechanism called \textit{social conventions}~\cite{busoniu2008comprehensive}, where the UAVs take turn to carry out an action. Each UAV is assigned with a specific ranking order. When considering the optimal joint action sets, the one with higher order will have priority to choose its action first, and let the subsequent one know its action. The other UAVs then can match their actions with respect to the selected action. To ensure collision avoidance, lower-ranking UAVs cannot take an action that will lead to the newly occupied states of higher-ranking UAVs in the system. By this, at a time step $k$, only one unique joint action will be agreed among the UAV's.

Defining the reward in MARL is another open problem due to the dynamic nature of the system~\cite{bucsoniu2010multi}. In this paper, the individual reward that each agent receives can be considered as the total number of cells it covered, minus the cells overlapping with other agents. However, a global team goal would help the team to accomplish the task quicker, and also speed up the learning process to converge faster~\cite{sadhu2017improving}. We define the global team's reward is a function $GR: \{S\} \times \{A\}\rightarrow \mathbb{R}$ that weights the entire team's joint state $S_{k}$ and joint action $A_{k}$ at time step $k$ in achieving (\ref{2.DiscreteObjective}). The agent only receives reward if the team's goal reached:
\begin{equation}\label{2.Reward}
\begin{aligned}
	\ GR(S_{k}, A_{k}) &= 
	\begin{cases}
	\  r, &  if  \ \sum_{i} f_{i}(S_{k}) \geq fb, \sum_{i} o_{i}(S_{k})\leq 0\\
    	 \ 0, &  otherwise. \\
    	\end{cases}
\end{aligned}
\end{equation}
\noindent
where $fb \in \mathbb{R}$ is an acceptable bound of the field being covered. During the course of learning, the state - action value function $Q_{i, k}(s_{i}, a_{i})$ for each agent $i$ at time $k$ can be iteratively updated as in Multi-Agent Q - learning algorithm, similar to those proposed in~\cite{sadhu2017improving, nowe2012game}:
\begin{equation}\label{2.UpdateMARL}
\begin{aligned}
	Q_{i, k+1}(S_{k}, A_{k})  &\leftarrow (1 - \alpha)Q_{i, k}(S_{k}, A_{k})+ \alpha[GR(S_{k}, A_{k})\\
					 &+ \gamma \underset{A' in \{A\}}{\max}Q_{i,k}(S_{k+1}, A') ], \\
\end{aligned}
\end{equation}
\noindent
where $0 < \alpha \leq 1$ is the learning rate, and $\gamma$ is the discount rate of the RL algorithm. The term $\underset{A' in \{A\}}{\max}Q_{i,k}(S_{k+1}, A')$ derived from (\ref{2.CE_pure}) at joint state $S_{k+1}$.

\subsection{Approximate Multi-Agent Q-learning}
In MARL, each agent updates its value function with respect to other agents' state and action, therefore the state and action variable dimensions can grow exponentially if we increase the number of agent in the system. This makes value function representation a challenge. Consider the value function $Q_{i, k+1}(S_{k}, A_{k})$ in (\ref{2.ValueFunctionJoint}), the space needed to store all the possible state - action pairs is $|\bold{S_{1}}|\cdot|\bold{S_{2}}|...\cdot|\bold{S_{m}}|\cdot|\bold{A_{1}}|\cdot|\bold{A_{2}}|...|\bold{A_{m}}| = |\bold{S_{i}}|^{m}|\bold{A_{i}}|^{m}$. 

Works have been proposed in the literature to tackle the problem: using graph theory~\cite{guestrin2002coordinated} to decompose the global Q-function into a local function concerning only a subset of the agents, reducing dimension of Q-table~\cite{zhang2017fmrq}, or eliminating other agents to reduce the space~\cite{borrajo2005reinforcement}. However, most previous approaches require additional step to reduce the space, that may place more pressure on the already-intense calculation time. In this work, we employ simple approximation techniques~\cite{geramifard2013tutorial}: Fixed Sparse Representation (FSR) and Radial Basis Function (RBF) to map the original Q to a parameter vector $\theta$ by using state and action - dependent basis functions $\phi:\{S\}\times\{A\} \rightarrow \mathbb{R}$:
\begin{equation}\label{2.QApproximated}
\begin{aligned}
	\hat{Q}_{i, k}(S_{k}, A_{k}) &= \sum_{l}\phi_{l}(S_{k}, A_{k})\theta_{i, l} = \phi^{T}(S_{k}, A_{k})\theta_{i},\\ 
\end{aligned}
\end{equation}
\noindent
The FSR scheme uses a column vector $\phi(S,A)$ of the size $D\cdot|\{A\}|$, where $D$ is the sum of \textit{dimensions} of the state space. For example, if the state space is a 3-D space: $X \times Y \times Z$, then $D = X+Y+Z$. Each element in $\phi$ is defined as follows:
\begin{equation}\label{2.BF}
\begin{aligned}
	\ \phi(x, y) &=
	\begin{cases}
	\  1, &  if  \ x = S_{k}, y = A_{k};\\
    	 \ 0, &  otherwise. \\
    	\end{cases}
\end{aligned}
\end{equation}
\noindent

In RBF scheme, we can use a column vector $\phi$ of $l\cdot|\{A\}|$ element, each can be calculated as:
\begin{equation}\label{2.RBF}
\begin{aligned}
	\ \phi(l, y) &=
	\begin{cases}
	\  e^{-\frac{S_{k}-c_{l}}{2\mu_{l}^{2}}}, &  if y = A_{k};\\
    	 \ 0, &  otherwise, \\
    	\end{cases}
\end{aligned}
\end{equation}
\noindent
where $c_{l}$ is the center and $\mu_{l}$ is the radius of $l$ pre-defined basis functions that have the shape of a Gaussian bell.

The $\phi(S,A) $ and $\theta_{i}$ in FSR and RBF schemes are column vectors of the size $D\cdot|\{A\}|$ and $l\cdot|\{A\}|$, respectively, which is much less than the space required in the original Q-value function. For instance, if we deploy 3 agents on a space of $7 \times 7 \times 4$, and each agent has $6$ actions, the original Q-table size would have $(7 \cdot 7 \cdot 7 \cdot 6)^{3} = 1.6\cdot 10^{9}$ numbers in it. Compare to the total space required for approximated parameter vectors in FSR scheme is $3\cdot(7 + 7 + 4) \cdot 6^{3}) = 3.8\cdot10^{3}$, and in RBF scheme is just $3\cdot8\cdot6 = 144$ numbers, the required space is hugely saved.

\begin{algorithm*}
\DontPrintSemicolon 
\KwIn{Learning parameters: Discount factor $\gamma$, learning rate $\alpha$, schedule $\{\epsilon^{k}\}$, number of step per episode $L$}
\KwIn{Basis Function vector $\phi(S,A)$, $\forall s_{i, 0} \in S_{i}$, $\forall a_{i, 0} \in A_{i}$}
Initialize $\theta_{i, 0} \leftarrow 0$, $ i = 1, ..., m$;\\
\For{$episode =  1, 2, ...$}{
	Randomly initialize state $s_{i, 0}$, $\forall i$\\
	\For{$k =  0, 1, 2,...$}{
		\For{$i =  0, 1, 2,..., m$}{
		Exchange information with other UAVs to obtain their state $s_{j,k}$ and parameters $\theta_{j}$, $j \neq i, j = 1...m$\\
		
		\begin{equation*}
		\begin{aligned}
		\ \ \pi(A_{k}) &=
		\begin{cases}
		\  \text{Find an optimal joint-action (strategy) by solving (\ref{2.CE_pure})}, & with \ probability \ 1 - \epsilon_{k}\\
    	 	\ \text{Take a random joint action}, & otherwise. \\
    		\end{cases}
		\end{aligned}
		\end{equation*}
		\noindent
		Decide unique joint action $A_{k}$, and take individual joint action according to \textit{social conventions} rule\\
		Receive other UAVs' new states ${s_{j,k+1}| j \neq i, j - 1,...,m}$ \\
		Observe global reward $r_{k+1} = GR(S_{k}, A_{k})$\\
		Update:
		\begin{equation*}
		\begin{aligned}
		\theta_{i, k+1}  &\leftarrow \theta_{i, k} + \alpha[GR(S_{k}, A_{k}) + \gamma \underset{A' in \{A\}}{\max} (\phi^{T}(S_{k+1},A') \theta_{i,k}) - (\phi^{T}(S_{k},A_{k})\theta_{i,k} ]\phi(S_{k}, A_{k}). \\
		\end{aligned}
		\end{equation*}
		\noindent
		}
	}
}
\KwOut{parameter vector $\theta_{i}, i = 1...m$ and policy $\pi$}
\caption{\sc Multi-Agent Approximated Equilibrium-based Q-Learning.}
\label{algo1}
\vspace{-3 pt}
\end{algorithm*}

After approximation, the update rule in (\ref{2.UpdateMARL}) for Q-function becomes the update rule for the parameter~\cite{busoniu2010reinforcement} set of each UAV $i$:
\begin{equation}\label{2.UpdateApproximatedMARL}
\begin{aligned}
	\theta_{i, k+1}  &\leftarrow \theta_{i, k} + \alpha[GR(S_{k}, A_{k}) \\
			&+ \gamma \underset{A' in \{A\}}{\max} (\phi^{T}(S_{k+1},A') \theta_{i,k}) \\
			&- (\phi^{T}(S_{k},A_{k})\theta_{i,k} ]\phi(S_{k}, A_{k}). \\
\end{aligned}
\end{equation}
\noindent
\begin{figure*}[htb!]
 \centering
	\subfloat[k = 1]{\includegraphics[width=0.25\textwidth]{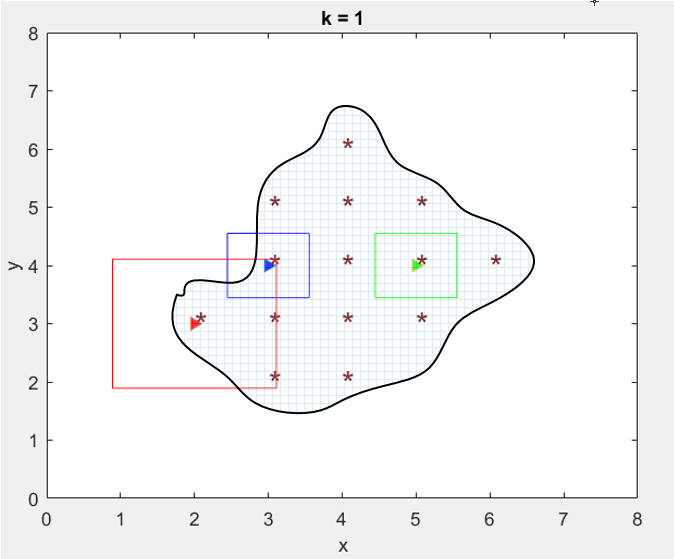}}
	\subfloat[k = 10]{\includegraphics[width=0.25\textwidth]{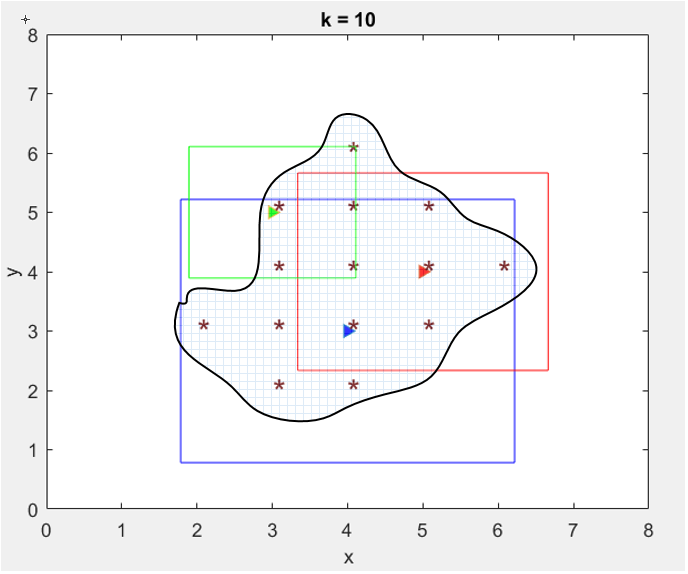}}
	\subfloat[k = 22]{\includegraphics[width=0.25\textwidth]{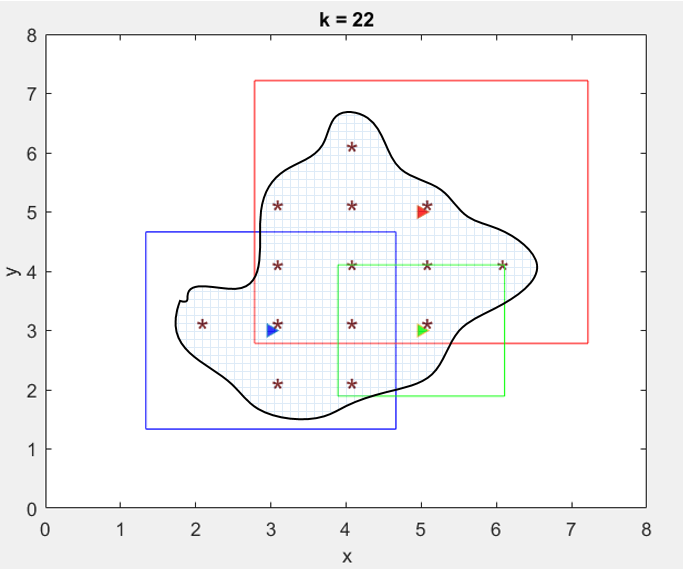}}
	\subfloat[k = 35]{\includegraphics[width=0.25\textwidth]{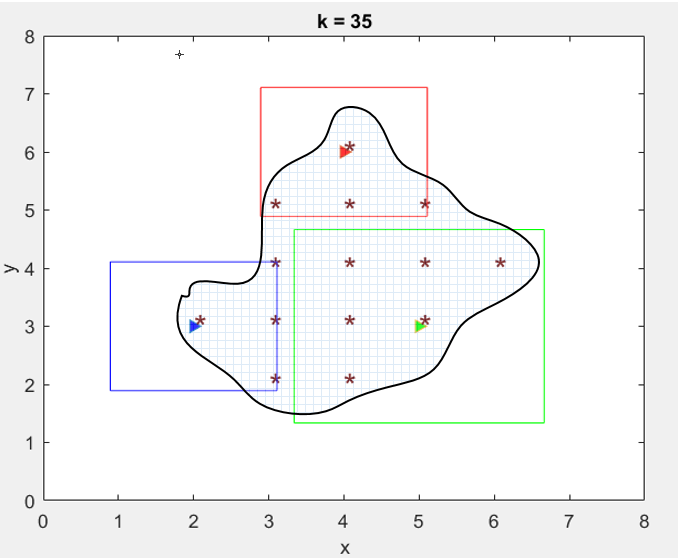}}
 \caption{2-D result showing the FOV of 3 UAVs collaborate in the last learning episode to provide a full coverage of the unknown field $F$ with discrete points denoted by $*$ mark, while avoiding overlapping others.}
   \label{SimFOV2D}
\vspace{-5 pt}
\end{figure*}
\begin{figure*}[htb!]
 \centering
	\subfloat[]{\includegraphics[width=0.25\textwidth]{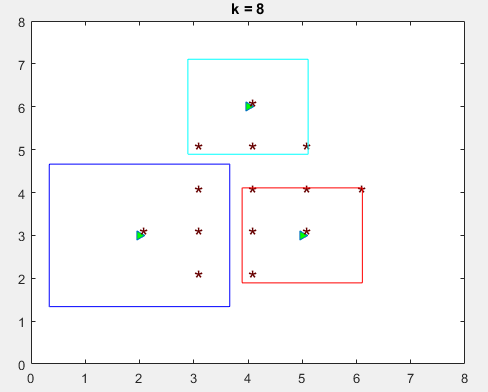}}
	\subfloat[]{\includegraphics[width=0.25\textwidth]{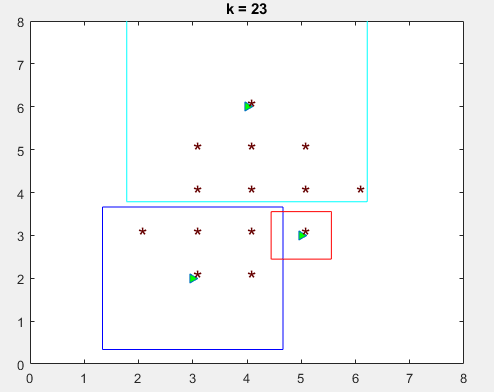}}
	\subfloat[]{\includegraphics[width=0.25\textwidth]{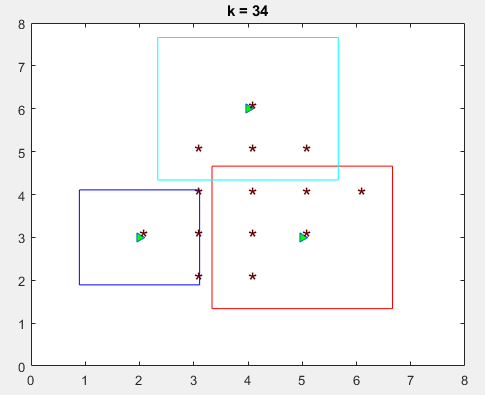}}
	\subfloat[]{\includegraphics[width=0.25\textwidth]{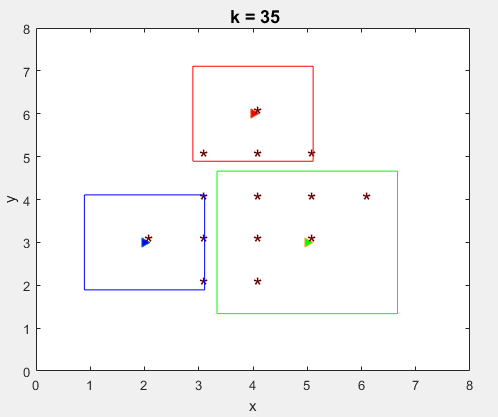}}
 \caption{Different optimal solutions show the configuration of the FOV of 3 UAVs with a full coverage and no discrete point ($*$ mark) overlapped.}
   \label{F.ConfigFOV}
\vspace{-5 pt}
\end{figure*}

\subsection{Algorithm}
We propose our learning process as Algorithm \ref{algo1}. The algorithm required  learning rate $\alpha$, discount factor $\gamma$, and a schedule $\{\epsilon^{k}\}$. The learning process is divided into episodes, with arbitrarily-initialized UAVs' states in each episode. We use a greedy policy $\pi$ with a big initial $\epsilon$ to increase the exploration actions in the early stages, but it will be diminished over time to focus on finding optimal joint action according to (\ref{2.CE_pure}). Each UAV will evaluate their performance based on a global reward function in (\ref{2.Reward}), and update the approximated value function of their states and action using the law (\ref{2.UpdateApproximatedMARL}) in a distributed manner.

\begin{figure}[htb!]
\centering
\includegraphics[width=0.8\columnwidth]{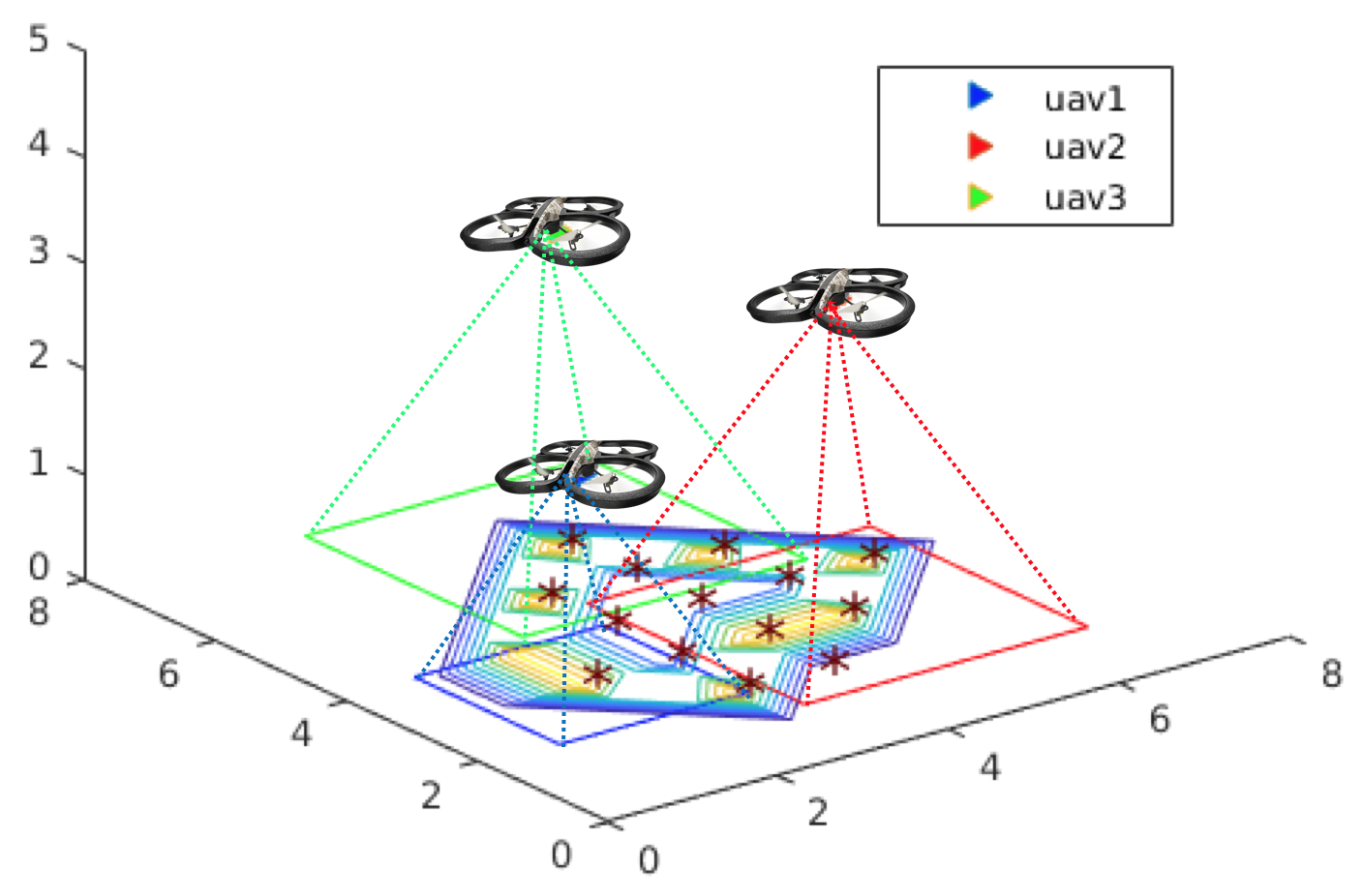}
  \caption{3-D representation of the UAV team covering the field.}
  \label{F.3D}
\vspace{-5 pt}
\end{figure}
\begin{figure}[htb!]
\centering
\includegraphics[width=1\columnwidth]{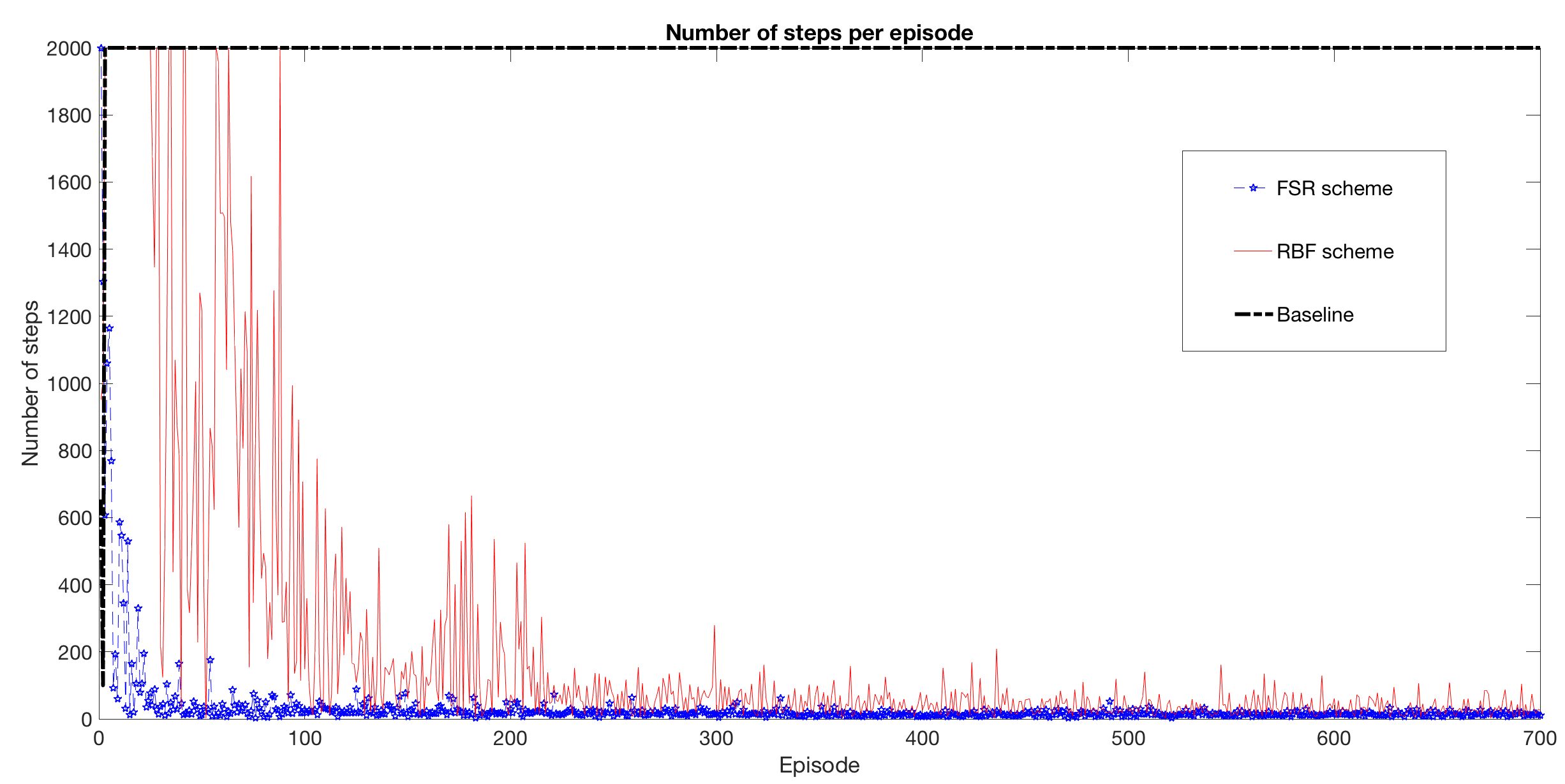}
  \caption{Number of steps the team UAV took over episodes to derive the optimal solution.}
  \label{F.SimSteps}
\vspace{-5pt}
\end{figure}
\begin{figure}[htb!]
\centering
\includegraphics[width=1\columnwidth]{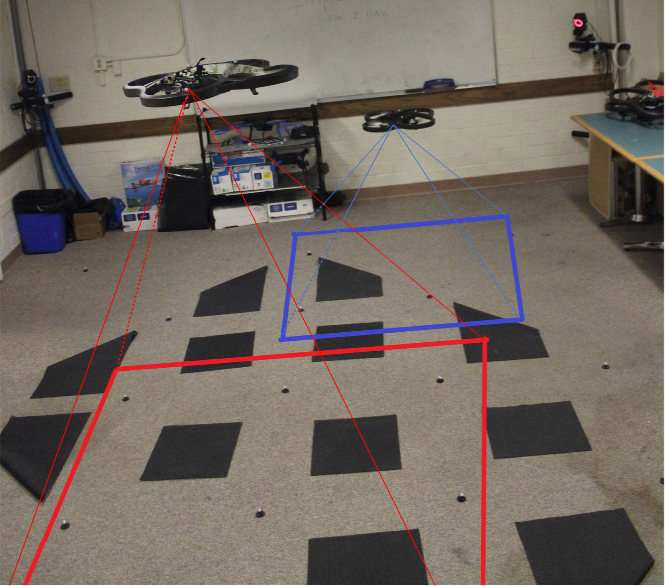}
  \caption{Physical implementation with 2 ARdrones. The UAVs cooperate to cover a field consists of black markers, while avoid overlapping each other.}
  \label{F.Real}
\vspace{-10 pt}
\end{figure}

\section{Experimental Results}\label{S.4}
\subsection{Simulation}
We set up a simulation on MATLAB environment to prove the effectiveness of our proposed algorithm. Consider our environment space as a $7 \times 7 \times 5$ discrete 3-D space, and a field of interest $F$ on a grid board with an unknown shape (Figure \ref{SimFOV2D}). The system has $m = 3$ UAVs, each UAV can take six possible actions to navigate: forward, backward, go left, go right, go up or go down. Each UAV in the team will have a positive reward $r = 0.1$ if the team covers the whole field $F$ with no overlapping, otherwise it receives $r = 0$. 

We implement the proposed algorithm \ref{algo1} with both approximation schemes: FSR and RBF, and compare their performance with a baseline algorithm. For the baseline algorithm, the agents seek to solve the problem by optimizing individual performance, that is to maximize their own coverage of the field $F$, and stay away from overlapping others to avoid a penalty of $-0.01$ for each overlapping square. For the proposed algorithm, both schemes use learning rate $\alpha = 0.1$, discount rate $\gamma = 0.9$, and $\epsilon = 0.9$ for the greedy policy which is diminished over time. To find CE for the agents in (\ref{2.CE_pure}), we utilize an optimization package for MATLAB from CVX~\cite{grant2008cvx}.

Our simulation on MATLAB shows that, in both FSR and RBF schemes after some training episodes the proposed algorithm allows UAV team to organize in several optimal configurations that fully cover the field while having no overlapping, while the baseline algorithm fails in most episodes. Figure \ref{SimFOV2D} shows how the UAVs coordinated to cover the field $F$ in the last learning episode in 2D. Figure \ref{F.ConfigFOV} shows a result of different solutions of the 3 UAV's FOV configuration with no overlapping. For a clearer view, Figure \ref{F.3D} shows the UAVs team and their FOVs in 3D environment in the last episode of the FSR scheme.

Figure \ref{F.SimSteps} shows the number of steps per episode the team took to converge to optimal solution. The baseline algorithm fails to converge, so it took maximum number of steps (2000), while the two schemes using proposed algorithm converged nicely. Interestingly, it took longer for the RBF scheme to converge, compare to the FSR scheme. It is likely due to the difference in accuracy of the approximation techniques, where RBF scheme has worse accuracy.

\subsection{Implementation}
In this section, we implement a lab-setting experiment for 2 UAVs to cover the field of interest $F$ with the similar specification as of the simulation, in an environment space as a $7 \times 7 \times 4$ discrete 3-D space. We use a quadrotor Parrot AR Drone 2.0, and the Motion Capture System from Motion Analysis~\cite{motionanalysis} to provide state estimation. The UAVs are controlled by a simple PD position controller~\cite{pham2018autonomous}.

We carried out the experiment using the FSR scheme, with similar parameters to the simulation, but now for only 2 UAVs. Each would have a positive reward $r = 0.1$ if the team covers the whole field $F$ with no overlapping, and $r = 0$ otherwise. The learning rate was $\alpha = 0.1$, and discount rate $\gamma = 0.9$, $\epsilon = 0.9$, which was diminished over time. Similar to the simulation result, the UAV team also accomplished the mission, with two UAVs coordinated to cover the whole field without overlapping each other, as showed in (Figure \ref{F.Real}).

\section{Conclusion}\label{S.5}

This paper proposed a MARL algorithm that can be applied to a team of UAVs that enable them to cooperatively learn to provide full coverage of an unknown field of interest, while minimizing the overlapping sections among their field of views. The complex dynamic of the joint-actions of the UAV team has been solved using game-theoretic correlated equilibrium. The challenge in huge dimensional state space has been also tackled with FSR and RBF approximation techniques that significantly reduce the space required to store the variables. We also provide our experimental results with both simulation and physical implementation to show that the UAVs can successfully learn to accomplish the task without the need of a mathematical model. In the future, we are interested in using Deep Learning to reduce computation time, especially in finding CE. We will also consider to work in more important application where the dynamic of the field presents, such as in wildfire monitoring.








\bibliography{ICRA2019_bibliography}
\bibliographystyle{IEEEtran}
\end{document}